\documentclass[10pt,twocolumn,letterpaper]{article}

\usepackage[pagenumbers]{cvpr} 

\usepackage{graphicx}
\usepackage{amsmath}
\usepackage{amssymb}
\usepackage{booktabs}
\usepackage{multirow}

\usepackage[pagebackref,breaklinks,colorlinks]{hyperref}

\usepackage[capitalize]{cleveref}
\crefname{section}{Sec.}{Secs.}
\Crefname{section}{Section}{Sections}
\Crefname{table}{Table}{Tables}
\crefname{table}{Tab.}{Tabs.}

\begin{document}

\title{QuAVF: Quality-aware Audio-Visual Fusion for Ego4D Talking to Me Challenge}

\author{Hsi-Che Lin$^1$ \quad Chien-Yi Wang$^{2}$ \quad 
Min-Hung Chen$^{2}$ \quad Szu-Wei Fu$^{2}$ \quad 
Yu-Chiang Frank Wang$^{1,2}$\\
\textsuperscript{1} National Taiwan University \quad
\textsuperscript{2} NVIDIA\\
{\tt\small hsichelin@gmail.com, \{chienyiw, minhungc, szuweif, frankwang\}@nvidia.com}
}
\maketitle

\begin{abstract}
This technical report describes our QuAVF@NTU-NVIDIA submission to the Ego4D Talking to Me (TTM) Challenge 2023. Based on the observation from the TTM task and the provided dataset, we propose to use two separate models to process the input videos and audio. By doing so, we can utilize all the labeled training data, including those without bounding box labels. Furthermore, we leverage the face quality score from a facial landmark prediction model for filtering noisy face input data. The face quality score is also employed in our proposed quality-aware fusion for integrating the results from two branches. With the simple architecture design, our model achieves $67.4\%$ mean average precision (mAP) on the test set, which ranks \textbf{first} on the leaderboard and outperforms the baseline method by a large margin. Code is available at: https://github.com/hsi-che-lin/Ego4D-QuAVF-TTM-CVPR23
\end{abstract}

\begin{figure*}[!ht]
\centering
\includegraphics[width=0.8\textwidth]{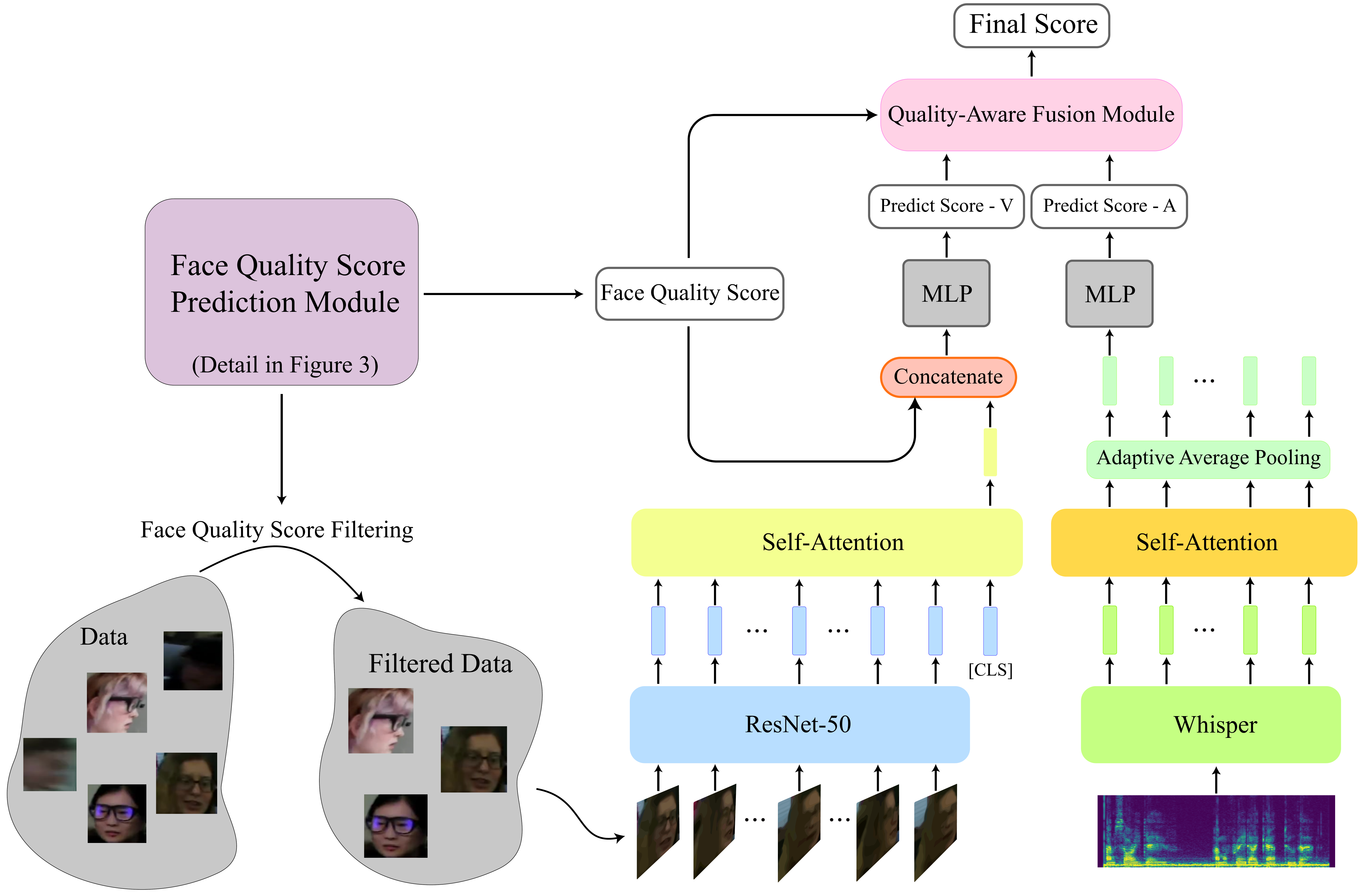}
\caption{\textbf{An illustration of our proposed Quality-aware Audio-Visual Fusion (QuAVF) framework.}}
\label{fig:pipeline}	
\end{figure*}

        

\section{Introduction}

Ego4D \cite{grauman2022ego4d} is a large-scale dataset introduced by Meta AI, specifically designed for the purpose of egocentric video understanding. Within the dataset, the Talking to Me (TTM) challenge focuses on the identification of social interactions in egocentric videos. Specifically, given a video and audio segment containing tracked faces of interest, the objective is to determine whether the person in each frame is talking to the camera wearer. This task holds significant importance for studying social interactions, and the understanding of egocentric social dynamics serves as a crucial element in various applications, including virtual assistants and social robots.

Drawing inspiration from the winning approach by Xue et al.~\cite{xue2023egocentric}, we attempt to fuse features from both modalities at an earlier stage. Therefore, in our initial approach, referred to as the Audio-Vision joint model (AV-joint), as shown in Figure~\ref{fig:avjoint}, we incorporate a fusion of vision and audio features immediately after the backbone network, prior to aggregating temporal information. The AV-joint model is trained by jointly optimizing the vision and audio branches.  However, despite employing a significantly larger backbone architecture (\textit{ResNet-50} \cite{he2016deep} and \textit{Whisper} \cite{radford2022robust}), the AV-joint model does not yield substantial performance improvements over the baseline model. Although our initial trial did not yield satisfactory results, a thorough analysis of the limited improvement led to several key observations that motivated our final approach. Firstly, as described in the original Ego4D paper \cite{grauman2022ego4d}, the determination of the TTM label is based on vocal activity, irrespective of whether the person is visible in the scene. Consequently, a significant portion of the training data lacks the corresponding bounding box label. (about 0.7M frames out of 1.7M frames with TTM label do not have bounding box label)

In our initial approach, we addressed the absence of bounding box labels by using zero padding. However, this approach can have adverse effects on the optimization process of the vision branch, as it may be trained on a large amount of non-realistic images. Additionally, since the visual and audio branches are trained jointly, the quality of the visual inputs can potentially impact the audio branch, particularly when fused at an early stage. Because the quality of the data is influenced by various factors, such as the methods employed to handle data without bounding box labels (e.g., zero padding), limitations of the hardware used to record egocentric videos, and potential inaccuracies in bounding box annotations, hence, improving data quality is not a straightforward task. One simple approach would be to discard data without bounding box labels, but this would significantly reduce the available data and waste audio activity annotations. To address these challenges, we explore disentangling the two modalities.

In our subsequent experiments, we discovered that using only the audio input resulted in superior performance compared to our initial AV-joint model(as shown in Table~\ref{tab:overall}.) This finding further reinforces our assumption that the quality of the visual data can impede the optimization process of the audio branch. As a result, in our final approaches, we employ separate models to process the audio and image modalities. For the audio branch, we leverage the powerful encoder from \textit{Whisper}~\cite{radford2022robust}, a robust speech recognition model, as we observed that the semantic information conveyed through conversations provides vital cues for this task. By disentangling the two modalities, the audio branch can fully utilize all the labels in the dataset, unaffected by variations in image quality. In the vision branch, we take steps to ensure data quality by incorporating an additional model that provides a quality score indicating the likelihood of a face appearing in images. This quality score is utilized to filter out inappropriate training data for the vision branch. Moreover, we discovered that employing a quantized quality score as supplementary information for the vision branch yields significant improvements to the model. Leveraging the same quality score, we introduce a quality-aware audio-visual fusion (QuAVF) approach. Our top-performing models achieved an accuracy of $71.2\%$ on the validation set and $67.4\%$ on the test set.

\vspace{-3mm}
\section{Approach}
In this section, we first describe the model architecture of our initial approach (AV-joint) in
detail. Then, we describe each component we designed in QuFAV. The final results and the
ablation study will be shown in the next section.

\subsection{Baseline Audio-Vision (AV) Joint Model}
\begin{figure}[t]
    \centering
    \includegraphics[width=0.32\textwidth]{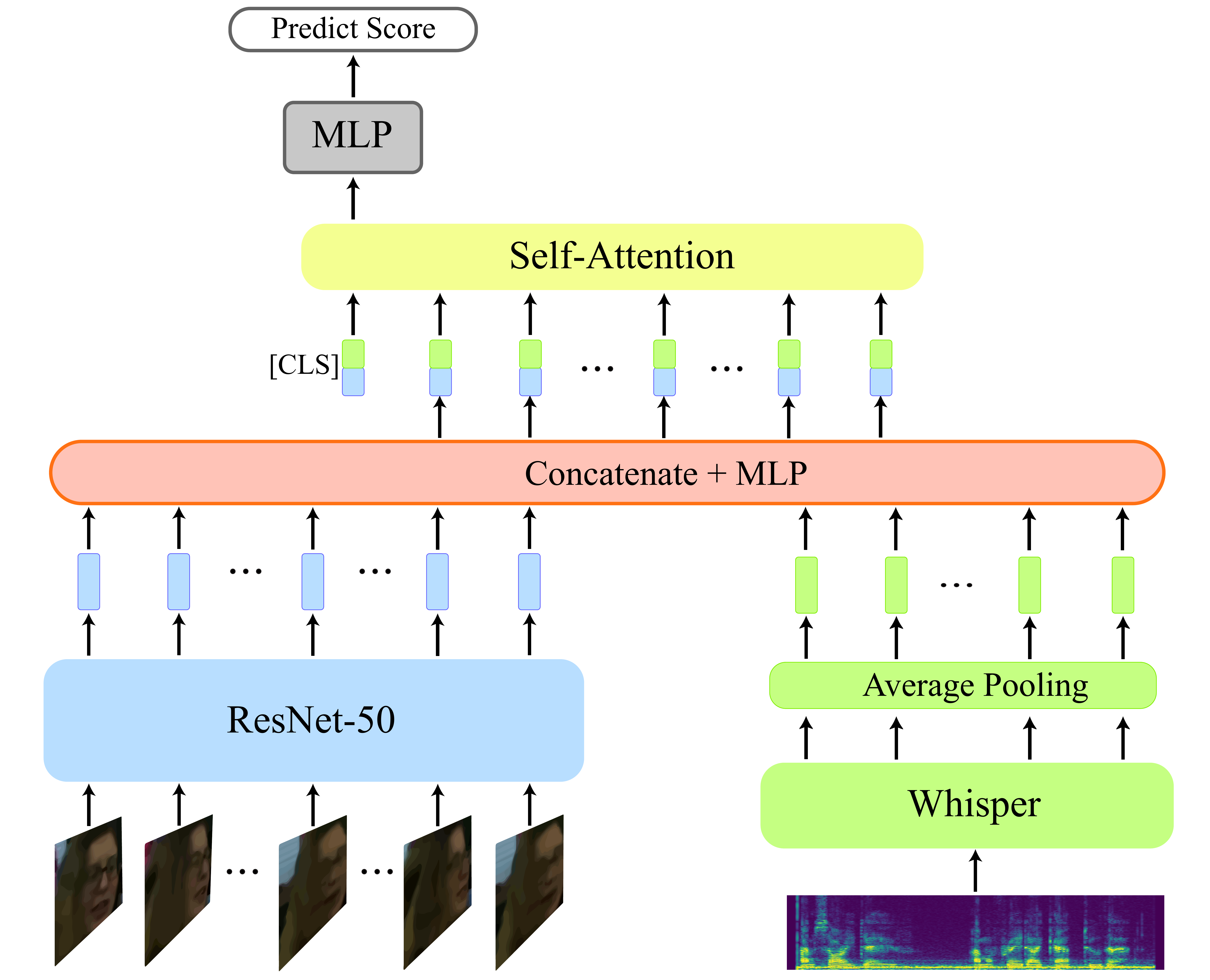}
    \caption{\small{\textbf{The baseline AV-joint model approach.}}}
    \label{fig:avjoint}
\end{figure}

\vspace{-5mm}
\noindent\paragraph{Model Architecture.}
In the AV-joint baseline model approach, the inputs are a sequence of 15 images sampled for the video
with a frame rate of 2, and the corresponding audio clip. We view the center image frame as our
target, and in each forward pass our model will predict the score of the target frame. We first extract
features from the audio and image sequence by the encoder of Whisper \cite{radford2022robust} and
the ResNet-50 \cite{he2016deep}, respectively. Since the length of the audio feature in the temporal
dimension will be longer than the vision feature (Whisper produces 50 features for 1-second audio,
while the video frame rate is 2) we apply average pooling to the audio features so that it has the same
length as the vision feature. We concatenate audio and vision features along the temporal dimension
followed by an MLP to fuse two modalities. Then, we append a [CLS] token and aggregate the
temporal information by self-attention layers. Finally, the prediction head is applied to the
corresponding output of the [CLS] token to produce the final prediction. The overall architecture of AV-joint is shown in Figure~\ref{fig:avjoint}.

\subsection{QuAVF}
\vspace{-1mm}
\subsubsection{Audio Branch}
\vspace{-7mm}
\noindent\paragraph{Model Architecture.}
For the audio branch, we use the encoder of Whisper-small \cite{radford2022robust}
as the backbone. We freeze the backbone and append a randomly
initialized self-attention layer to refine the temporal information. Given an audio clip, we follow
the approach described in \cite{radford2022robust}, pad the clip to 30 seconds, and transform it
into an 80-channel log-magnitude Mel spectrogram. Since the length of audio input may not match
the number of image frames, we apply an adaptive average pooling and a prediction head on the output
of the self-attention layer to obtain the final logits for each image frame. The model architecture of audio branch can be found in Figure~\ref{fig:pipeline}

\vspace{-5mm}
\noindent\paragraph{Augmentation.}
We have tried to add Gaussian noise or crop the input audio for data augmentation. With probability, we sample a Gaussian noise with a  predetermined range of signal-to-noise ratio (SNR). Also, with probability, we randomly crop the input audio into a shorter clip whose length is also uniformly sampled from a predetermined range.

\vspace{-4mm}
\subsubsection{Vision Branch}
\vspace{-8mm}
\noindent\paragraph{Model Architecture.}
As for the vision branch, we choose a \textit{ResNet50} \cite{he2016deep} pre-trained on ImageNet
as the backbone. To predict the label of a specific frame, we choose 7 frames with a frame rate of $2$ before
and after it ($15$ frames or $7.5$ seconds in total), and feed them to our backbone independently. 
We then apply two randomly initialized self-attention layers to the extracted features (together with an
additional learnable [CLS] token) to aggregate the information on the temporal
dimension. Finally, a prediction head is applied to the [CLS] token to obtain the result logits.
The model architecture of vision branch can be seen in Figure~\ref{fig:pipeline}

\begin{figure}[t]
    \centering
    \includegraphics[width=0.41\textwidth]{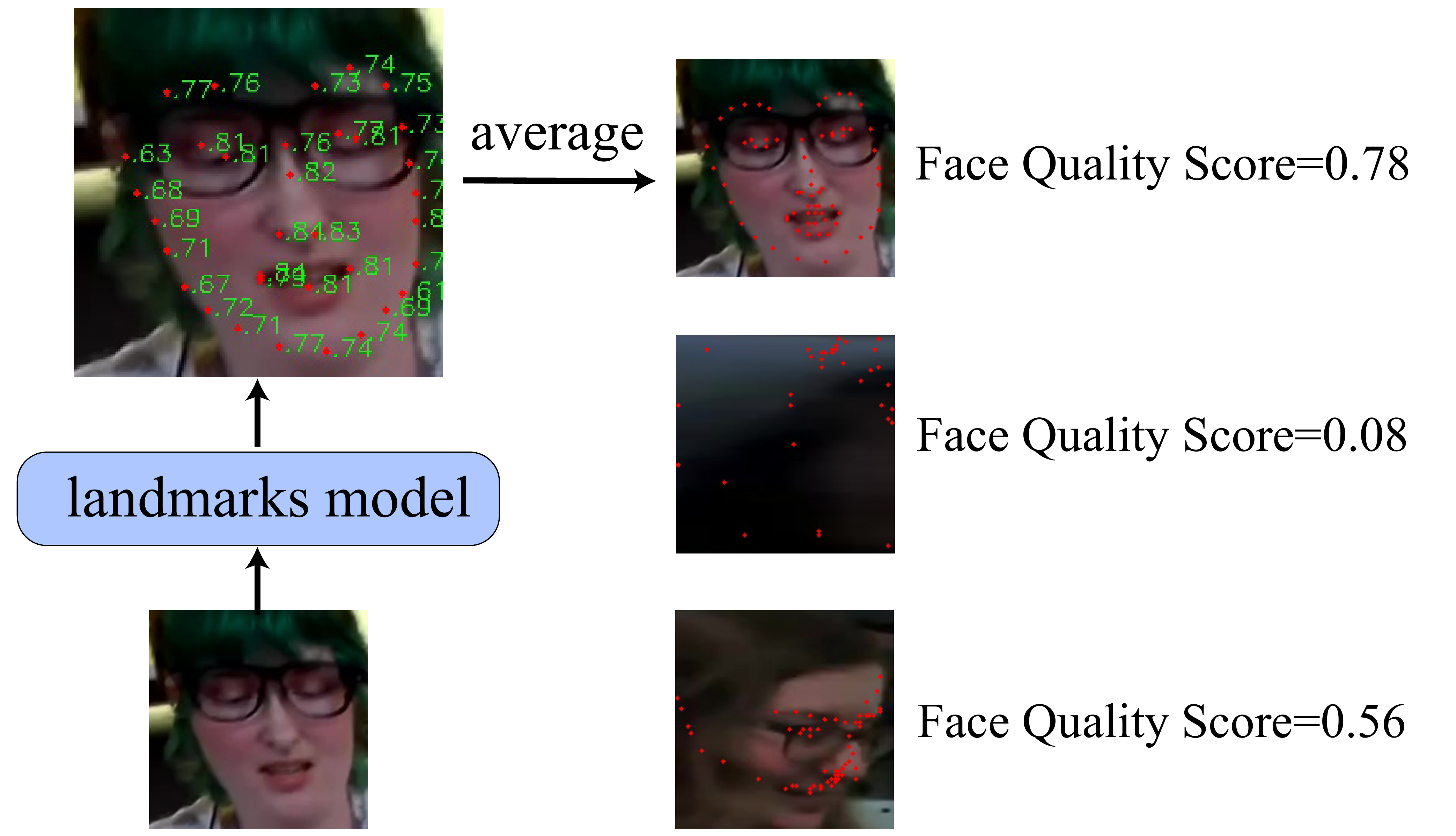}
    \caption{\small{\textbf{An illustration of face quality score computation with the facial landmark model~\cite{bulat2017far}.}}}
    \label{fig:FaceQualityScore}
\end{figure}

\vspace{-4mm}
\noindent\paragraph{Data Filtering.}
In addition to disentangling the vision and audio modalities, we've also tried to improve the quality
of training data for the vision branch. To that end, we apply the facial landmarks prediction model~\cite{bulat2017far} on the bounding box region of training data and average the confidence scores
of all the landmark points (Figure~\ref{fig:FaceQualityScore}). We treat the resulting score as the face quality score for that image, which represents how likely there is a face appearing in that region. The face quality score of a training sample is defined as the average of face quality score over all the included frames. The data with a score lower than a threshold will be discarded to increase the data quality.

\vspace{-5mm}
\noindent\paragraph{Auxiliary Face Quality Score Feature.}
The face quality scores obtained from the landmarks estimation model are not only used to filter the
data but also used as an input feature. We experiment with two different settings. The first one is to apply
a linear transformation on the scalar and concatenate the output feature with the final [CLS] token before the prediction head. The second one is to quantize the score first. The result of quantization
will be a one-hot vector showing which level of magnitude the score falls into. We then apply the
transformation, concatenation, and prediction head on this vector.

\vspace{-3mm}
\subsubsection{Quality-Aware Fusion Module}
Since we use two independent model to process the audio and images separately, to make our final decision consider information from both modalities, we need an additional fusion module. To that end, we introduce a quality-aware fusion module, which considers the face quality score and fuse the prediction scores from two branches. The design is very simple; that is, we simply compute the weighted sum of score from each branch with the weight of the vision branch set as the face quality score (the weight of the audio branch is then $(1-$face quality score$)$). The overall pipeline of our QuAVF is shown in Figure~\ref{fig:pipeline}.

\section{Experiment}

\subsection{Experimental Setup}

We evaluate the proposed AV-joint, QuAVF, and the two branches in QuAVF on the validation and test data. For each model, we choose the setting that has best performance on the validation data, and we apply the same moving average post-process with widow size 25 to the raw prediction. We report the result together with some previous approaches in Table~\ref{tab:overall}.

\begin{table}[h]
    \begin{minipage}[t]{1\linewidth}
        \centering\setlength{\tabcolsep}{4pt}
        \footnotesize
        \renewcommand{\arraystretch}{1.2}
        \caption{\small{Results of Talking to Me (TTM) challenge.}}
        \vspace{-3mm}
        \resizebox{1.0\linewidth}{!}
        {
            \begin{tabular}{c|cc|cc}
                \hline
                \multirow{2}{*}{method} & \multicolumn{2}{c|}{Validation}    & \multicolumn{2}{c}{Test}  \\
                                        & Accuracy  & mAP                   & Accuracy  & mAP \\
                \hline
                \hline
                Random Guess \cite{grauman2022ego4d}            & & & 47.41 & 50.16 \\
                \hline
                ResNet-18 Bi-LSTM \cite{grauman2022ego4d}       & & & 49.75 & 55.06 \\
                \hline
                EgoTask Translation \cite{xue2023egocentric}    & & & 55.93 & 57.51 \\
                \hline
                Baseline AV joint        & 58.1  & 59.5  & 55.66 & 57.05 \\
                \hline
                Audio-only    & 71.7  & 70.1  & \textbf{57.77} & \textbf{67.39} \\
                Vision-only   & 64.0  & 67.2  & 54.80 & 56.17 \\
                \hline
                QuAVF           & \textbf{71.8}  & \textbf{71.2}  & 57.05 & 65.83 \\
                \hline
            \end{tabular}
        }
        \vspace{2mm}
        \label{tab:overall}
    \end{minipage}
    \vspace{-3mm}
\end{table}

For the ablation study, we evaluate our model with different settings on the validation data. If not mentioned otherwise, we freeze the backbone model owing to the hardware constraints. We report the resulting accuracy and mAP on the validation data in Table~\ref{tab:avjoint} ~\ref{tab:audioonly} ~\ref{tab:visiononly} ~\ref{tab:postprocess}.

\vspace{-1mm}
\subsection{Results}

From Table~\ref{tab:overall}, we can see that our QuAVF outperforms the previous winning approach in terms of both accuracy and mAP. Also, we can see the effect of disentangling vision and audio branch as well as the quality-aware fusion.

We have tried different backbone model in our initial approach AV-joint, and the results are shown in in Table~\ref{tab:avjoint}. We find that the choice of the backbone is not very decisive, which suggests that there should be other reasons behind the limited improvement of AV-joint over the baseline model. As a result, we started to explore the characteristics of input data, which motivates our quality filtering method.

\begin{table}[t]
    \begin{minipage}[t]{1\linewidth}
        \centering\setlength{\tabcolsep}{4pt}
        \footnotesize
        \renewcommand{\arraystretch}{1.2}
        \caption{\small{Results of baseline AV joint model on validation data}}
        \vspace{-3mm}
        \resizebox{0.8\linewidth}{!}
        {
            \begin{tabular}{c|c|cc}
                \hline
                method  
                    & backbone                          & Accuracy  & mAP   \\
                \hline
                \hline
                \multirow{2}{*}{AV joint}
                    & ResNet-50 Whsiper                 & \textbf{53.6}      & \textbf{59.5}  \\
                    & AV-HuBERT \cite{shi2022avhubert}  & 53.4      & 58.2  \\
                \hline
            \end{tabular}
        }
        \vspace{2mm}
        \label{tab:avjoint}
    \end{minipage}
    \vspace{-3mm}
\end{table}

For the audio branch, we experiment with the augmentation methods. The result is presented in Table~\ref{tab:audioonly}. We have tried different range of SNR of the input. However, we find that noise augmentation does not have much effect. In contrast, randomly crop the input audio can consistently improve the performance. We experiment with different probability to apply cropping as well as the minimum length after cropping.

\begin{table}[h]
    \begin{minipage}[t]{1\linewidth}
        \centering\setlength{\tabcolsep}{4pt}
        \footnotesize
        \renewcommand{\arraystretch}{1.2}
        \caption{\small{Results of audio branch model on validation data}}
        \vspace{-3mm}
        \resizebox{1.0\linewidth}{!}
        {
            \begin{tabular}{c|c|c|cc}
                \hline
                method  
                    & noise aug. (SNR)  & random crop   & Accuracy  & mAP   \\
                \hline
                \hline
                \multirow{6}{*}{Audio only}
                    &                   &               & 72.8      & 68.5  \\
                    & min=3, max=20     &               & 68        & 64.8  \\
                    & min=10, max=20    &               & 70.9      & 68.1  \\
                    &                   & p=0.5, min=3  & \textbf{72}        & 69.5  \\
                    &                   & p=0.9, min=3  & 71.7      & \textbf{70.1}  \\
                \hline
            \end{tabular}
        }
        \vspace{2mm}
        \label{tab:audioonly}
    \end{minipage}
    \vspace{-3mm}
\end{table}

We experiment with different filter threshold when training the vision branch. The result can be seen in Table~\ref{tab:visiononly}. Also, from the table we can see that the face quality scores indeed provide useful information especially when the scores are quantized. By training the whole model including the backbone model, we obtain our best model for vision branch. The moving average post process method can consistently improve the results as shown in Table~\ref{tab:postprocess}

\begin{table}[h]
    \begin{minipage}[t]{1\linewidth}
        \centering\setlength{\tabcolsep}{4pt}
        \footnotesize
        \renewcommand{\arraystretch}{1.2}
        \caption{\small{Results of vision branch model on validation data}}
        \vspace{-3mm}
        \resizebox{1.0\linewidth}{!}
        {
            \begin{tabular}{c|c|c|cc}
                \hline
                method  
                                & filter    & face quality score    & Accuracy  & mAP   \\
                \hline
                \hline
                \multirow{6}{*}{Vision branch}
                                &           &                       & 53        & 54    \\
                                & $>0.5$    &                       & 51.3      & 55.9  \\
                                & $>0.3$    &                       & 57.2      & 62    \\
                                & $>0.3$    & scalar                & 57        & 62    \\
                                & $>0.3$    & quantized             & 63        & 65    \\
                (fine-tune)     & $>0.3$    & quantized             & \textbf{64}        & \textbf{67.2}  \\
                \hline
            \end{tabular}
        }
        \vspace{2mm}
        \label{tab:visiononly}
    \end{minipage}
    \vspace{-3mm}
\end{table}

\begin{table}[t]
    \begin{minipage}[t]{1\linewidth}
        \centering\setlength{\tabcolsep}{4pt}
        \footnotesize
        \renewcommand{\arraystretch}{1.2}
        \caption{\small{Effect of moving average post-process on validation data}}
        \vspace{-3mm}
        \resizebox{0.5\linewidth}{!}
        {
            \begin{tabular}{c|c|c}
                \hline
                \multirow{2}{*}{method branch}  & \multicolumn{2}{c}{mAP}   \\
                                & w/out & with  \\
                \hline
                \hline
                av-joint        & 59.5  & \textbf{59.9}  \\
                audio branch    & 70.1  & \textbf{70.4}  \\
                vision branch   & 67.2  & \textbf{67.4}  \\
                \hline
            \end{tabular}
        }
        \vspace{2mm}
        \label{tab:postprocess}
    \end{minipage}
    \vspace{-3mm}
\end{table}

\section{Discussion}


\vspace{-1mm}
\subsection{Performance Gap between Validation and Test}
\vspace{-2mm}
Although our QuAVF method obtains the best $71.2\%$ mAP on the validation data, it does not perform
equally well on the test data. From the results, we can see that making the audio branch independent of
vision input improves the model on both validation and test data. Hence, we suspect that this
performance gap comes from the vision branch, especially from the distribution difference between
validation and test quality score. However, future works are needed to find out the real reason.


\begin{figure}[t]
    \centering
    \begin{subfigure}{20mm}
        \includegraphics[width=20mm]{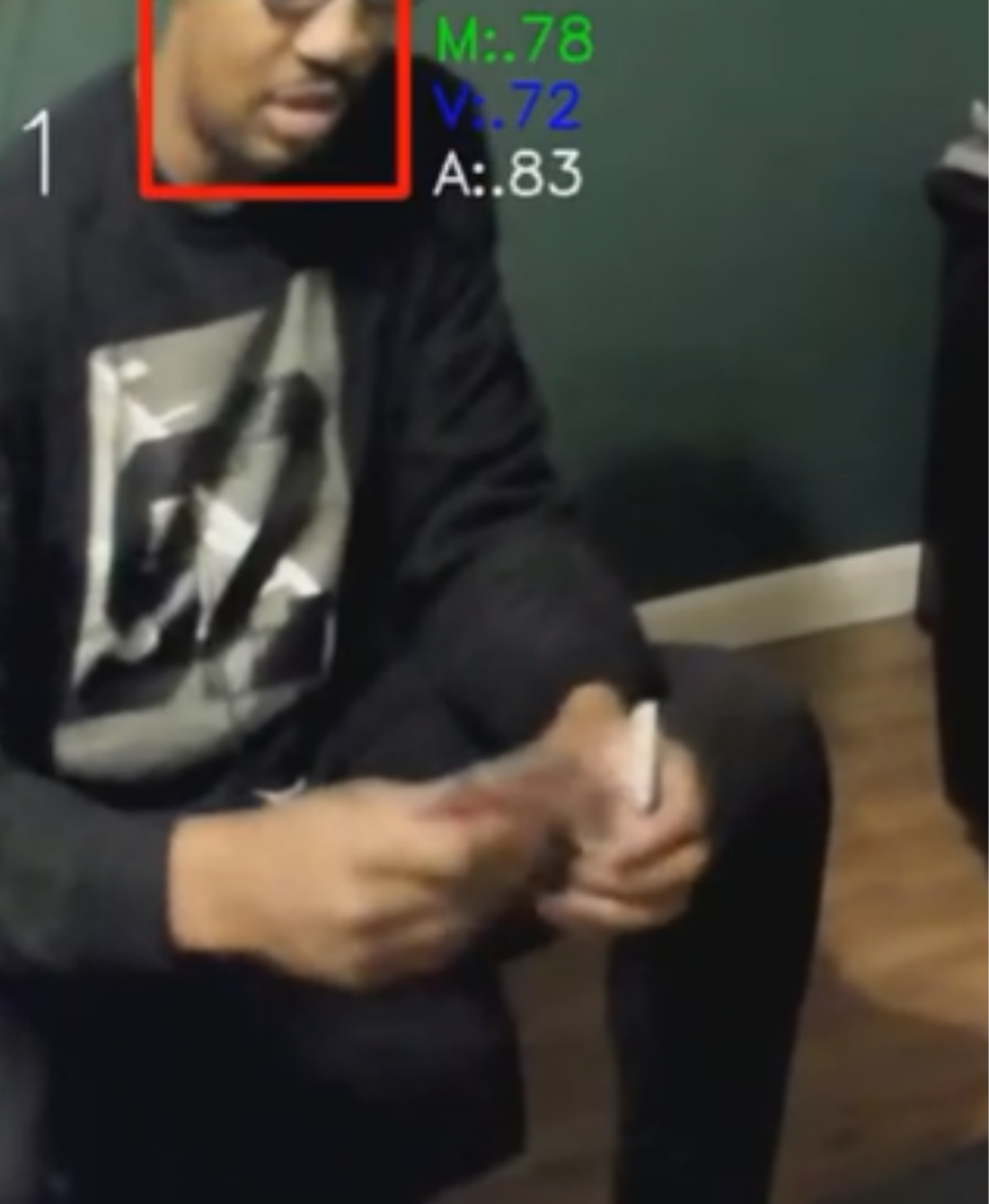}
    \end{subfigure}
    \begin{subfigure}{20mm}
        \includegraphics[width=20mm]{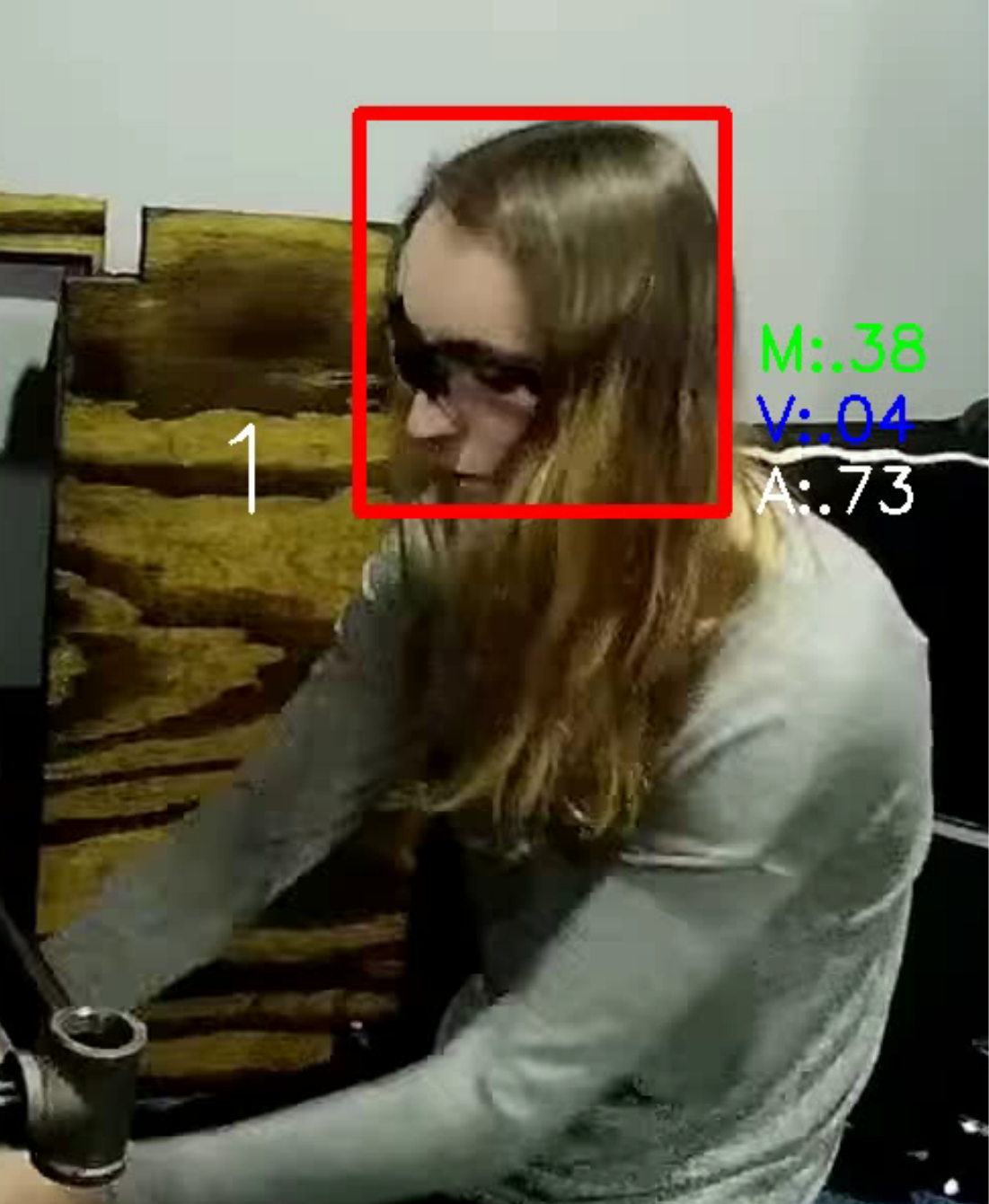}
    \end{subfigure}
    \begin{subfigure}{20mm}
        \includegraphics[width=20mm]{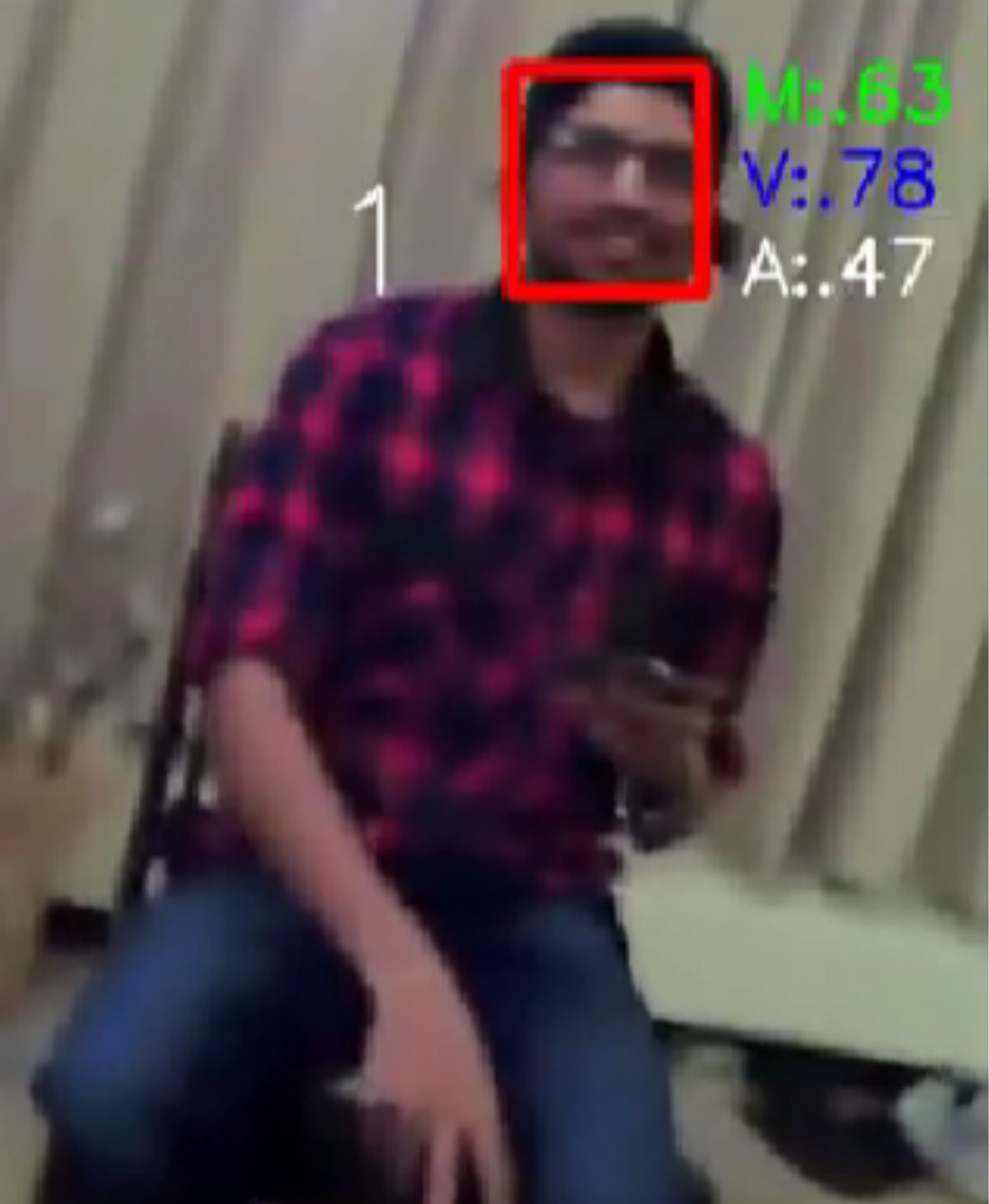}
    \end{subfigure}
    \begin{subfigure}{20mm}
        \includegraphics[width=20mm]{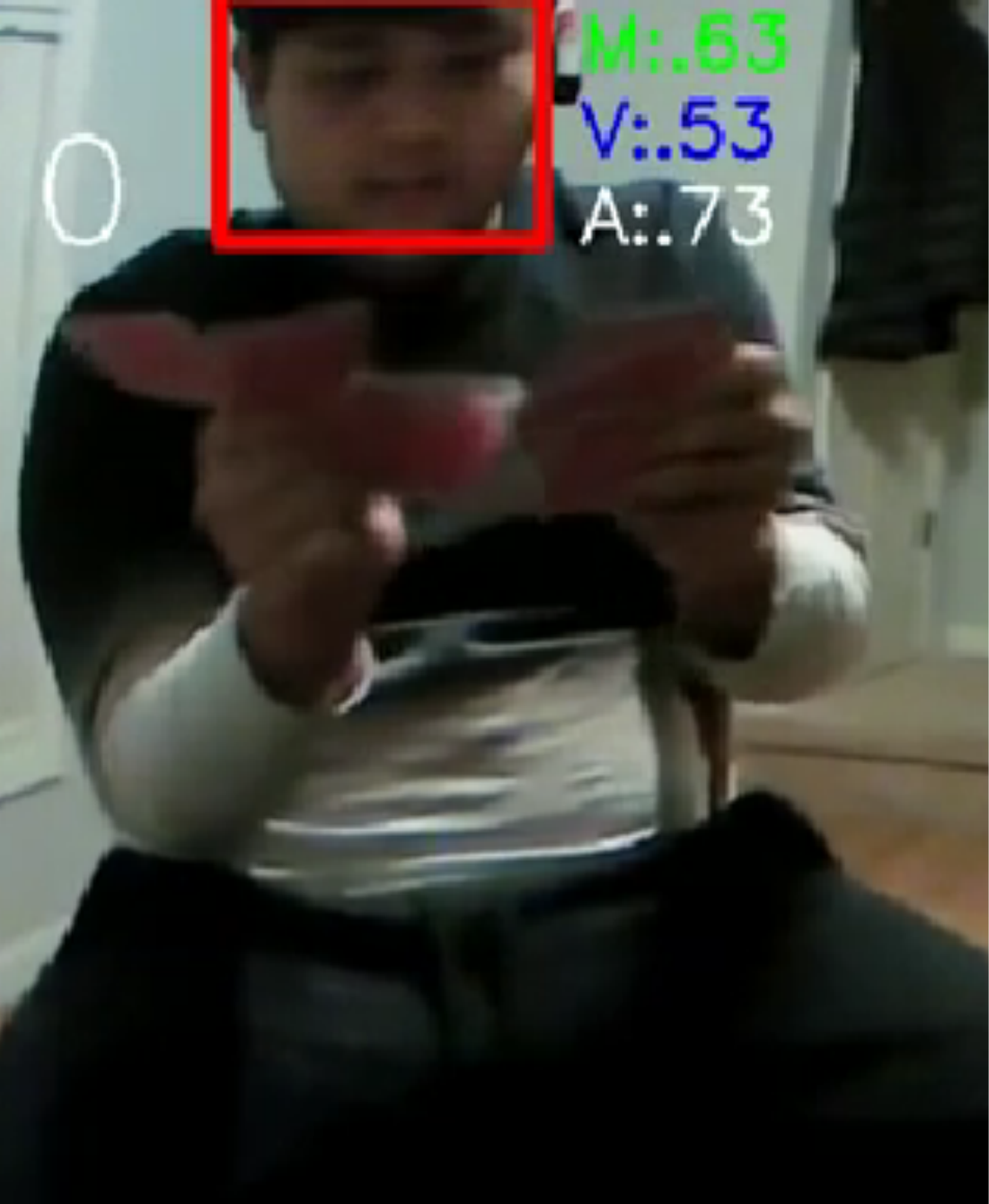}
    \end{subfigure}
    \caption{\small{\textbf{Illustrating positive and negative examples}}}
    \label{fig:PosNegExamples}
\end{figure}

\subsection{Positive and Negative Examples}
In Figure~\ref{fig:PosNegExamples}, we give positive and negative examples on the validation set. The white number on the bottom left corner is the ground truth TTM label. On the bottom right corner are the prediction score of QuAVF, vision branch, and audio branch, respectively (top to bottom).

{\small
\bibliographystyle{ieee_fullname}
\bibliography{egbib}
}

\end{document}